\documentclass[preprint]{article}

\usepackage[final,nonatbib]{neurips_2020}

\newcommand{\makeExtended}[0]{1}
\newcommand{\contentBasePath}[0]{content/}
\newcommand{\visBasePath}[0]{\contentBasePath vis/}
\usepackage[utf8]{inputenc} 
\usepackage[english]{babel}
\usepackage{amsmath}
\usepackage{amssymb}
\usepackage{graphicx} 
\usepackage{xspace} 
\usepackage{todonotes}
\usepackage{subfig} 
\usepackage[normalem]{ulem}
\usepackage{yfonts}
\usepackage[backend=biber, style=numeric-comp, sorting=nyt, maxbibnames=10, natbib=true]{biblatex}
\usepackage{wrapfig}
\usepackage[export]{adjustbox}
\usepackage{hyperref}
\hypersetup{breaklinks=true}

\newboolean{vc_is_included}
\gdef\GITAbrHash{b785d11}%
\gdef\VCDateTEX{2021/07/26}%
}{
	\setboolean{vc_is_included}{false}
	
	\gdef\GITAbrHash{}

	\gdef\VCDateTEX{}

}

\newboolean{article_extended}
\ifthenelse{\isundefined{\makeExtended}}{
	\setboolean{article_extended}{false}}{
	\setboolean{article_extended}{true}}

\newboolean{article_dense_spacing}
\ifthenelse{\isundefined{\makeDenseSpacing}}{
	\setboolean{article_dense_spacing}{false}}{
	\setboolean{article_dense_spacing}{true}}

\newlength{\articleEnvSpaceTop}
\newlength{\articleEnvSpaceTopSmall}
\newlength{\articleEnvSpaceBottom}
\ifthenelse{\boolean{article_dense_spacing}}{
	\setlength{\articleEnvSpaceTop}{-2cm}
	\setlength{\articleEnvSpaceTopSmall}{-1cm}
}{
	\setlength{\articleEnvSpaceTop}{-1cm}
	\setlength{\articleEnvSpaceTopSmall}{0cm}
	\setlength{\articleEnvSpaceBottom}{-1cm}
}

\newcommand{\articleTitle}[0]{Experiments on Properties of Hidden Structures of Sparse Neural Networks}

\newcommand{\articleAuthor}[0]{Julian Stier, Harshil Darji, Michael Granitzer}

\newcommand{\articleEmail}[0]{julian.stier@uni-passau.de, darji01@ads.uni-passau.de}

\title{\articleTitle}

\author{\articleAuthor\\
	\texttt{\articleEmail}
	\ifthenelse{\boolean{vc_is_included}}{\tiny\protect\\\tiny\VCDateTEX~$\sim$~\GITAbrHash}{~}}

\begin{document}

\maketitle

\begin{abstract}
	Sparsity in the structure of Neural Networks can lead to less energy consumption, less memory usage, faster computation times on convenient hardware, and automated machine learning.
If sparsity gives rise to certain kinds of structure, it can explain automatically obtained features during learning.

We provide insights into experiments in which we show how sparsity can be achieved through prior initialization, pruning, and during learning, and answer questions on the relationship between the structure of Neural Networks and their performance.
This includes the first work of inducing priors from network theory into Recurrent Neural Networks and an architectural performance prediction during a Neural Architecture Search.
Within our experiments, we show how magnitude class blinded pruning achieves 97.5\% on MNIST with 80\% compression and re-training, which is 0.5 points more than without compression, that magnitude class uniform pruning is significantly inferior to it and how a genetic search enhanced with performance prediction achieves 82.4\% on CIFAR10.
Further, performance prediction for Recurrent Networks learning the Reber grammar shows an $R^2$ of up to 0.81 given only structural information.


\end{abstract}

\maketitle

\section{Introduction}\label{sec:introduction}
Understanding the structure of deep neural networks promises advances across many open problems such as energy-efficient hardware, computation times, and domain-specific performance improvements.
The structure is coupled with sparsity on different levels of the neural architecture, and if there is no sparsity, then there is also no structure:
a single hidden layered neural network is capable of universal approximation \cite{kidger2020universal}, but as soon as there exists a deeper structure, there naturally occurs sparsity.

Clearly, the structure between the input domain and the first hidden layer is tightly coupled with the structure within the data -- correlations between the underlying random variables such as the spatial correlation of images or correlation in windows of time series data.
In theory and with perfectly fitting functions, that should be all there is, but in practice, neural architectures got deeper and deeper, and hidden structures seem to have an effect when neural networks are not just measured by their goodness of fit but also, e.g., on hardware efficiency or robustness \cite{benamor2021robustness}.
Assuming such hidden structures exist for the better, we wonder how we can automatically find them, how they can be controlled during learning, and whether we can exploit given knowledge about them.

We give our definition for sparse neural networks and show experiments on automatic methods to obtain hidden structures:
pruning, neural architecture search, and prior initialization.
With structural performance prediction, we also show experiments on exploiting structural information to speed up neural architecture search methods.

Our \textbf{contributions} comprise a pytorch tool called \textit{deepstruct}\footnote{\url{http://github.com/innvariant/deepstruct}} which provides models and tools for Sparse Neural Networks, a \textit{genetic neural architecture search} enhanced with structural performance prediction, a \textit{comparison of magnitude-based pruning} on feed-forward and recurrent networks, an \textbf{original} correlation analysis on \textit{recurrent networks} with different biologically plausible \textbf{structural priors} from social network theory, and \textit{performance prediction} results on these recurrent networks.
Details on the experiments and code for reproducibility can be found at \href{https://github.com/innvariant/sparsity-experiments-2021}{github.com/innvariant/sparsity-experiments-2021}.

\section{Sparse Neural Networks}\label{sec:sparse}
Sparse Neural Networks (SNNs) are deep neural networks $f$ with a low proportion of connectivity $\xi(f)$ with respect to all possible connections.

\paragraph{Sparsity}
Given a vector $x \in \mathbb{R}^d$ with $d\in\mathbb{N}$, its sparsity is $\xi(x) = \frac{||x||_0}{d} = \frac{1}{d}\cdot\sum_{i=0}^d |x_i|^0$, given the cardinality function $||\cdot||_0$ (of which 0 refers to the case of $p = 0$ of a $\mathcal{L}_p$ norm) and the size of the vector.
Density is defined as its complement with $1-\xi(x)$.
The definition extends naturally to tensors and simply provides the proportion of non-zero elements in a tensor compared to the total number of its elements.
A tensor can be considered as sparse as soon as its sparsity is below a given threshold value, e.g., $\xi(x) < 0.5$ -- as soon as more than 50\% of its elements are zero.

What is the \textbf{motivation} for sparsity at all?
First, more sparsity implies a lower number of parameters which is desirable if the approximation and generalization capabilities are not heavily affected.
In theory, it also implies a lower number of computations.
From a technical perspective, sparse structures could lead to specialized hardware.
Further, sparsity means that there is space for compression that can affect the overall model memory footprint.
Memory requirements are an important aspect for limited capacity devices such as in mobile deployment.
In the feature transformation layers, sparsity explains data dependencies and provides room for explainability.

\paragraph{Neural Networks}
A neural network is a function composed of non-linear transformation layers $\sigma(Wx+B)$ extended with transformations for skip-layer connections such that $z^l = \sum_{s=1}^{l-1} W^{s\rightarrow l}\cdot a^s + B^l$ with $a^l = \sigma(z^l)$ being the activation of layer $l$ with $\sigma$ being e.g. \textit{tanh} or \textit{max(x, 0)}.
$W^{s\rightarrow l}$ describes the weights from layer $l-1$ to $l$ for a network with $l\in\{1,\dots,L\}$.
The input to the function $a^0$ is $x \in \mathbb{R}^{d_x}$ from the input domain.
Consecutive sizes of weight matrices $W$ need to be aligned and define the layer sizes.
The final weight matrices $W^{s\rightarrow L}$ map to the output domain $\mathbb{R}^{d_y}$ with $B^L \in \mathbb{R}^{d_y}$.

Given the weights of a neural network $f$ as a set of grouped vectors $W$, we overload $\xi$ such that we obtain the sparsity of a neural network $\xi(f) = \frac{1}{|W|}\sum_{x\in W} \xi(x)$.
A \textbf{Sparse Neural Network} is a neural network $f$ with low sparsity, e.g. $\xi(f) < 0.5$.
The set of grouped vectors could, e.g., be all neurons with their weights from all possible incoming connections.

\ifthenelse{\boolean{article_extended}}{
	Sparse Neural Networks (SNN) naturally provide a directed acyclic graph $G = (V,E)$ associated with them.
	Neurons defined in $W^{s\rightarrow l}$ translate to vertices, and weighted connections from a neuron in layer $s$ to a neuron in layer $l$ translate to edges between the according vertices.
	The resulting graph structure contains reduced information about the original model.
	Technically, the graph can be reflected in an SNN with masks $M^{s\rightarrow l}$ associated in a multiplicative manner through the Hadamard product $\odot$ with each $W^{s\rightarrow l}$, and we get
	\[
		z_l = \sum\limits_{s=1}^{l-1} \left(W^{s\rightarrow l}\odot M^{s\rightarrow l}\right)\cdot a^s + B^l
	\]
	This mask can be either obtained through learned weights by, e.g., pruning or regularization or induced by prior design.

	All deep feed-forward neural networks are sparse by design as they lack residual connections.
	Even ResNets and DenseNets contain sparsity as they are employing convolutions which are sparse by design \cite{he2016deep,zhang2018residual}.
	Techniques such as poolings compute order statistics over their input and thus must be either excluded or can be understood as a structural singularity.
}{}

\paragraph{Sparse Recurrent Neural Networks (SRNN)}
Recurrent Neural Networks additionally have recurrent connections which unfold over time.
These recurrent connections are initialized as hidden states, $h$.
At any sequence $t$, $h_t = \sigma (Wx_t + Uh_{t-1} + b_h), t \in \{1, ..., T\}$ with $W$ and $U$ being input-to-hidden weigths and hidden-to-hidden weights, respectively.
We refer to $x_t$ as the input at sequence $t$ and $h_{t-1}$ as the hidden state value from the previous step.

Similar to SNNs, SRNNs also consists of extended non-linear transformation layers with skip-layers such that $h_t^l = \sum_{s=1}^{l-1} W^{s\rightarrow l}h_t^{s} + Uh_{t-1}^l + b_h^l$ with $t \in \{1,...,T\}$.

\begin{wrapfigure}{r}{0.5\textwidth}
	\vspace{-1em}
	\centering
	\includegraphics[width=0.5\textwidth]{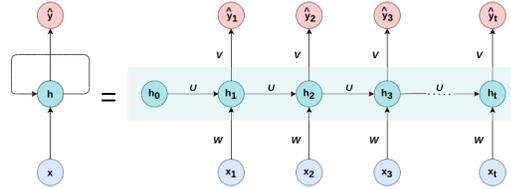}
	\caption{
		A simple \textbf{Recurrent Neural Network} unrolled with $t$ sequences. Here, $W$ represents \textit{input-to-hidden weights}, $U$ represents \textit{hidden-to-hidden weights}, and $V$ represents \textit{hidden-to-output weights}.
	}
	\label{fig:rnn_simple}
	\vspace{-2em}
\end{wrapfigure}

In SRNNs, directed acyclic graphs are not associated with recurrent connections, only with consecutive transformation layers.
Therefore, to reflect a graph in an SRNN, only input-to-hidden weights $W^{s\rightarrow l}$ are multiplied with masks $M^{s\rightarrow l}$ through the Hadamard product $\odot$ such that $h_t^l$ can be formulated as:

\[
	\sum\limits_{s=1}^{l-1} \left(W^{s\rightarrow l}\odot  M^{s\rightarrow l}\right)h_t^{s} + Uh_{t-1}^l + b_h^l
\]

\paragraph{Achieving Sparsity}
Sparsity refers to a structural property of Neural Networks which can be desirable due to reasons such as model size or hardware acceleration.
There exist multiple ways to achieve sparsity, e.g., through regularization, pruning, constraints, or by prior initialization.

\textit{Regularization} affects the optimization objective such that not only a target loss but also a parameter norm is minimized.
As such, regularization takes effect during training and can force weights to be of small magnitude.
Under sufficient conditions, e.g., with an L1-norm and rectified linear units as activation functions, sparsity in the trained network can be achieved in an end-to-end fashion during learning.

\textit{Pruning} refers to removing elements of the network outside of the end-to-end training phase.
Based on a selection criterion such as the magnitude of a weight, one or multiple weights can be set to zero.
A pruning scheme decides on what sets the criterion is applied or how often the pruning is repeated.
Sparsity is enforced based on this selection criterion and pruning scheme.

\textit{Prior design} is a constraint on the overall search space of the optimization procedure.
More generally, prior structure to a neural network is restricting the hypothesis space of all possible obtainable functions during learning to a smaller space.
Convolutions as feed-forward neural networks with local spatial restrictions can be understood as such a prior design.

\paragraph{Pruning Neural Networks}\label{sec:pruning}
Pruning is a top-down method to derive a model from an originally larger one by iteratively removing elements of the neural network.
The motivation to prune is manifold:
1) finding high-performing network structures can be faster in comparison to other search methods such as grid- or random-search, 2) pruning can improve metrics such as error, generalization, fault tolerance, or robustness or 3) reduce computational costs in terms of speed, memory requirements and energy efficiency or 4) support the interpretation of neural networks.

Pruning consists of a selection method and a strategy.
The selection method decides which elements to choose based on a criteria, e.g., the magnitude of a weight.
The strategy applies the selection method repeatedly on a model until some stopping criterion is reached, e.g., a certain number of iterations are conducted.

One-shot or single pruning refers to applying the pruning method once.
After pruning, often a certain number of re-training cycles are conducted.
Fixed-size pruning refers to selecting a fixed number of elements based on the ranking obtained through the pruning selection method.
In each step, the same number of elements are removed.
Relative or percentage pruning refers to selecting a percentage of remaining elements to be pruned.
This results in fewer numbers to be removed in sudden decays of performance.
Bucket pruning holds a bucket value which is filled by, e.g., the weight magnitude or the saliency measure of the pruning selection method, and as many elements as the bucket can hold are removed per step.

A naive method for pruning is the random selection of $k$ components.
Differences can be made by defining the granularity, e.g., whether to prune weights, neurons, or even channels or layers.
Random pruning often serves as a baseline for pruning methods to show their general effectiveness, and it has been shown in various articles that most magnitude- and error-based methods outperform their random baseline, see Figure \ref{fig:ffn-pruning}.
Usually, models drop in performance after pruning but recover within a re-training phase of few epochs.

For magnitude-based pruning, good explanations can be found in \cite{marchisio2018prunet, han2015learning} and in recent surveys such as \cite{liang2021pruning, gale2019state}.
Class-blinded selects weights based on their magnitude regardless of their class, i.e., their layer, class-uniform selects the same amount of weights from each class, and class-distributed selects elements in proportion to the standard deviation of weight magnitudes in the respected class.

\paragraph{Prior Design}\label{sec:prior-design}
Restricting the search space of the neural network architecture fosters faster convergence, and domain-specific improved performance can be achieved.
Convolutions with kernels applied over spatially related inputs are a good example for such a prior.
Similarly, realizations as the MLP-Mixer \cite{tolstikhin2021mlp} show that even multi-layer perceptrons with additional imposed structure and final poolings can achieve state-of-the-art performance.

\section{Related Work}\label{sec:related-work}

\paragraph{Pruning}
\ifthenelse{\boolean{article_extended}}{
	Pruning dates as far back as 1988 \cite{sietsma1988neural} in which Sietsma et al. provided the first error-based approach to prune hidden units from a neural network.
	Notable contributions of this time have been Optimal Brain Damage and its modification Optimal Brain Surgeon \cite{lecun1990optimal,hassibi1993optimal} besides techniques such as Skeletonization \cite{mozer1989skeletonization} which introduced estimating the sensitivity of the error, fault-tolerance \cite{segee1991fault} and improved generalization \cite{thodberg1991improving}.
	Magnitude-based approaches followed with weight-elimination \cite{weigend1991generalization, weigend1991back}.
	There have also been several specializations of pruning, such as \cite{sankar1991optimal}.
	Reed \cite{reed1993pruning} provides a well-known survey on pruning algorithms up until then.
}{}

Recent work on pruning has been conducted by Han et al. \cite{han2015learning, han2015learning} using different magnitude-based pruning methods for deep neural networks in the context of compression or by Dong et al. \cite{dong2017learning}.
A survey by Liang et al. \cite{liang2021pruning} provides extensive insights into pruning and quantization for deep neural networks.
In 2018, authors of the Lottery Ticket Hypothesis reported on finding sparse subnetworks after iteratively pruning, re-setting to the original weight initialization, and training it from scratch to a comparable performance \cite{frankle2018lottery}.
We collected over 300 articles on \href{https://github.com/JulianStier/nn-pruning}{pruning} just up until 2019.

First pruning experiments on Recurrent Network Network were performed by Lee et al. \cite{giles1994pruning}.
Han et al. \cite{han2017adaptive} proposed recurrent self-organising neural networks, adding or pruning the hidden neurons based on their competitiveness during training.
A Baidu research group \cite{narang2017exploring} could reduce the network size by 8$\times$ while managing the near-original performance.
In 2019, Zhang et al. \cite{zhang2019one} proposed one-shot pruning for Recurrent Neural Networks using the recurrent Jacobian spectrum.
Using this technique, the authors confirmed that their network, even with 95\% sparsity, performed better than fully dense networks.

\paragraph{Sparse Training}
Besides $L_1$- or $L_2$-regularizations which can lead to real-zeros with appropriate activation functions such as ReLUs, there are also training methods outside the optimization objective to enforce sparsity such as 
Sparse Evolutionary Training (SET) by Mocanu et al. \cite{mocanu2018scalable}, Dynamic Sparse Reparamterization (DSR) \cite{mostafa2019parameter} or the Rigged Lottery (RigL) \cite{evci2020rigging}.
For RNNs, there exists Selfish sparse RNN training \cite{liu2021selfish}.

\paragraph{Neural Architecture Search} On Neural Architecture Search (NAS), there are two notable recent surveys by Elsken et al. \cite{elsken2019neural} and Wistuba et al. \cite{wistuba2019survey} providing an overview and dividing NAS into the definition of a \textbf{search space}, a \textbf{search strategy} over this space and the \textbf{performance estimation} strategy.
Differentiable architecture search \cite{liu2018darts} is a notable method for finding sparse neural networks on a high-level graph based search space by allowing to choose among paths in a categorical and differentiable manner.
While there exist hundreds of variations in the definition of search spaces and methods, the field recently came up with benchmarks and comparable metrics \cite{ying2019bench}.

\paragraph{Structural Performance Prediction} Structural Performance Prediction refers to using structural features of a neural network to predict a performance estimate without any or only partial training.
Such performance prediction was already conducted by Baker et al. \cite{baker2018accelerating} on two structurally simple features, namely the total number of weights and the number of layers.
But they mostly focused on prediction based on hyperparameters and time-series information.
Klein et al. also did performance prediction based on time-series information.
They conducted ``learning curve prediction with Bayesian Neural Networks'' \cite{klein2016learning}.
In \cite{stier2019structural} more extensive graph properties of randomly induced structures were used to predict the performance of neural networks for image classification.
A related work on a ``genetic programming approach to design convolutional neural network architectures'' \cite{wendlinger2021evofficient} included an acceleration study for accuracy prediction based on path depth, breadth-first-search level width, layer out height and channels, and connection type counts.
The performance prediction during a NAS yields a 1.69$\times$ speed-up.
Similar to \cite{stier2019structural} in \cite{benamor2021robustness} we used structural properties to predict the robustness of recurrent neural networks.

\section{Experiments}\label{sec:experiments}
We conducted four experiments:
First, pruning feed-forward neural networks to investigate the effect of different pruning methods, namely random pruning, magnitude class-blinded, magnitude class-uniform, magnitude class-distributed, and Optimal Brain Damage \cite{lecun1990optimal}.
Second, pruning recurrent neural networks to investigate whether we observe similar compression rates and to have a baseline comparison for recurrent models in the subsequent experiment.
Third, inducing random graphs as structural priors into recurrent neural networks, based on the biological motivation that biological neural networks are also connected like small-world networks \cite{hilgetag2016brain}.
And fourth, conducting a genetic neural architecture search with architectural performance prediction\ifthenelse{\boolean{article_extended}}{ -- a complex search space with a strategy that can be accelerated when having information about properties of sparse genotypes.}{.}


%
%
\subsection{Pruning Feed-Forward Networks}


On \texttt{MNIST} \cite{lecun1998gradient} we used two different feed-forward architectures with rectified linear units with 100 and 300-100 neurons in the hidden layers, trained up to 200 epochs with cross-entropy, a batch size of 64, and a learning rate of 0.01 and 0.0001.
With the five pruning methods, we conducted several repeated experiments with iterative fixed-size pruning or iterative relative pruning of the number of weights.

We found that magnitude class blinded pruning clearly outperforms the other methods and Optimal Brain Damage, surprisingly, performs nearly the same although it uses second-order derivative information for pruning.
Pruning in general can dramatically reduce overfitting in the examined network and can even outperform other regularization techniques.

\begin{figure}[tbh]
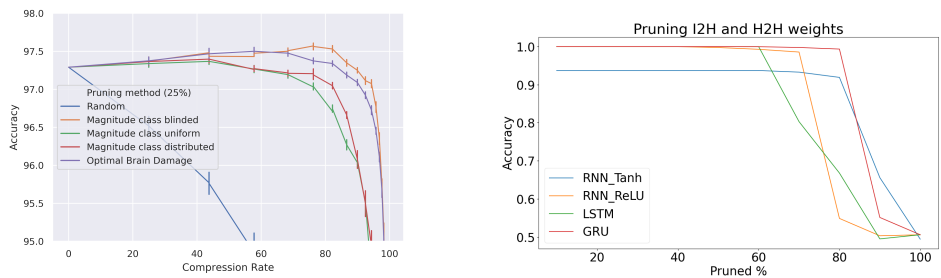

	\vspace{-1em}
	\centering
	\subfloat[Five pruning methods on feed-forward networks.
		Choosing weight-magnitude over random selections clearly has advantages.
		Optimal Brain Damage is expensive and worth in a low-parameter regime.\label{fig:ffn-pruning}]{
		\includegraphics[width=0.43\textwidth]{\visBasePath comp-magnitude-san.png}
	}
	~
	\subfloat[Pruning both input-to-hidden and hidden-to-hidden weights on a recurrent neural network.\label{fig:both_prune_performance}]{
		\includegraphics[width=0.49\textwidth]{\visBasePath prune_performance.png}
	}
	\caption{Pruning performance in feed-forward networks (Figure \ref{fig:ffn-pruning}) and recurrent networks (Figure \ref{fig:both_prune_performance}).}
	\vspace{-2em}
\end{figure}

%
%
\subsection{Pruning Recurrent Networks}

\begin{figure}[tbh]
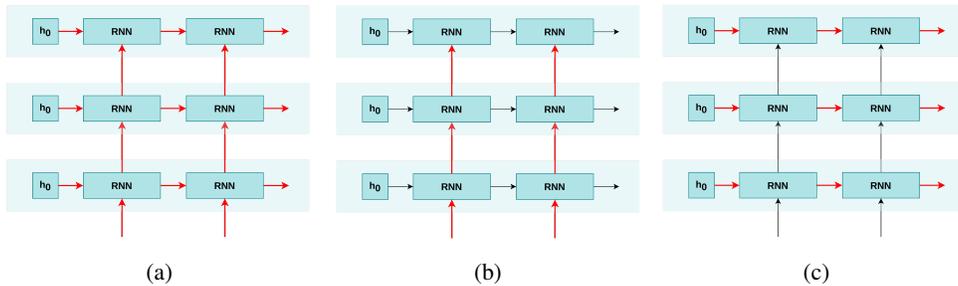

  	\vspace{\articleEnvSpaceTop}
	\centering
	\subfloat[\label{fig:both_prune}]{
		\includegraphics[width=0.3\textwidth]{\visBasePath both_prune.png}
	}
	\subfloat[\label{fig:i2h_prune}]{
		\includegraphics[width=0.3\textwidth]{\visBasePath i2h_prune.png}
	}
	\subfloat[\label{fig:h2h_prune}]{
		\includegraphics[width=0.3\textwidth]{\visBasePath h2h_prune.png}
	}
  	\caption{
  		The red right arrow (\textcolor{red}{$\longrightarrow$}) resembles pruning of hidden-to-hidden weights, the red up-arrow (\textcolor{red}{$\big\uparrow$}) pruning of input-to-hidden weights.
  		Figure \ref{fig:both_prune} shows pruning both, Figure \ref{fig:i2h_prune} only i2h weights, and Figure \ref{fig:h2h_prune} only h2h weights.
  	}
  	\label{fig:pruning}
  	\vspace{-1em}
\end{figure}

In this experiment, we prune input-to-hidden``i2h" and hidden-to-hidden ``h2h" weights individually and simultaneously as depicted in Figure \ref{fig:pruning} on a pre-trained base recurrent model for the \texttt{Reber} grammar \cite{reber1976implicit}, trained for 50 epochs.
\ifthenelse{\boolean{article_extended}}{
	\begin{wrapfigure}{r}{0.5\textwidth}
		\vspace{-2em}
		\centering
		\includegraphics[width=0.5\textwidth]{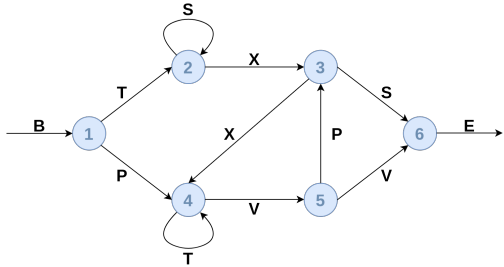}
		\caption{
			Flow diagram to generate \texttt{Reber} grammar sequences.
		}
		\label{fig:reber}
		\vspace{-1.5em}
	\end{wrapfigure}

	One of the first uses of \texttt{Reber} grammar was in the paper that introduced the LSTM \cite{hochreiter1997long}. 
	A true \texttt{Reber} grammar sequence follows the flow-chart shown in Figure \ref{fig:reber}. 
	For example, \textbf{BPTVPXTTVVE} is a true \texttt{Reber} grammar sequence, while \textbf{BTTVPXTVPSE} will be the false one.
}{}
The base recurrent model consists of an embedding layer that accepts input in the form of ASCII values of each character in the input \texttt{Reber} sequence.
Three recurrent layers of 50 neurons follow the embedding layer and a linear layer predicts the final scores.
TanH and ReLU are used as non-linearities.
The models are trained with a learning rate of 0.001 and a batch size of 32.

From the \texttt{Reber} grammar, we generated 25000 sequences, out of which 12500 are true \texttt{Reber} grammar sequences, and the remaining are false.
The dataset resembles a binary classification task in which a model has to predict whether a sequence is in the \texttt{Reber} grammar or not.
Logically, a baseline performance from random guessing is accuracy of 50\%.
We then split this dataset into a train-test split of 75\%-25\%, with 18750 sequences in the training set and 6250 sequences in the test set.

The threshold based on which we prune weights is calculated based on the percent of weights to prune.
Therefore, for example, to prune $p=10\%$ of weights for a given layer, the threshold is the 10th percentile of all the \textbf{absolute} weights in that layer.
In our experiment, we go from percent $p=10$ to $100$ while incrementing $p$ by $10$ after each round.

We considered LSTM, GRU, and vanilla RNNs as architectures for comparison.
Base models were trained separately for each to get the base performance.
Then, we pruned i2h and h2h weights; simultaneously and individually. 
Based on these results, we can identify the effect pruning has on the performance of RNNs and the amount of re-training required to regain the original performance.

The base models for RNN\_ReLU, LSTM, and GRU achieve perfect accuracies of 100\% on the test set within the first two epochs.
RNN\_TanH achieved 90\% after six epochs and showed a drop in accuracy between epoch three and five down to 50\%, which we observed over multiple repetitions of the experiment.


Pruning both i2h and h2h weights simultaneously, about 80\% of weights in RNN\_Tanh, 70\% of weights in RNN\_ReLU, 60\% of weights in LSTM, and 80\% of weights in GRU can be safely reduced as can be observed in Figure \ref{fig:both_prune_performance}.
After pruning above the safe threshold, we re-trained each pruned model and found that it only takes one epoch to regain the original performance.
Pruning 100\% weights, the model never recovers.

\begin{figure}[tbh]
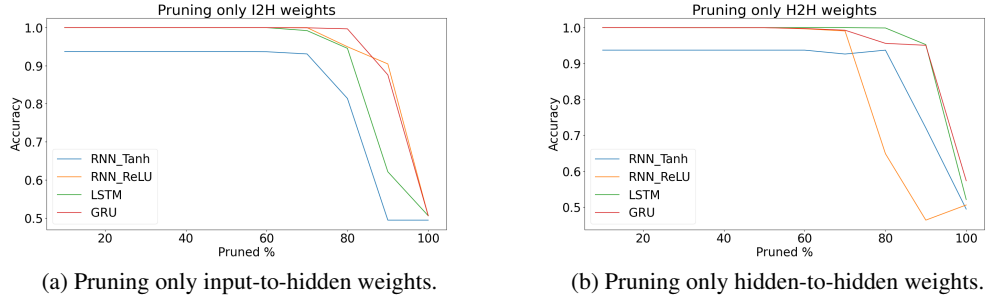

  	\vspace{\articleEnvSpaceTop}
	\centering
	\subfloat[Pruning only input-to-hidden weights.\label{fig:i2h_prune_performance}]{
		\includegraphics[width=0.49\textwidth]{\visBasePath i2h_prune_performance.png}
	}
	~
	\subfloat[Pruning only hidden-to-hidden weights.\label{fig:h2h_prune_performance}]{
		\includegraphics[width=0.49\textwidth]{\visBasePath h2h_prune_performance.png}
	}
  
  	\caption{
  		Accuracies of RNN\_Tanh, RNN\_ReLU, LSTM, and GRU after applying iterative magnitude percentage pruning on a common base model.
  	}
  	\label{fig:prune_performance}
  	\vspace{-2em}
\end{figure}

Pruning only i2h weights, results showed that we safely prune about 70\% for RNN\_Tanh, RNN\_ReLU, and LSTM.
For GRU, we prune 80\% of i2h weights without noticing a significant reduction in performance, see Figure \ref{fig:i2h_prune_performance}.
As in the case of pruning both i2h and h2h weights simultaneously, our pruned model still recovers only after one re-training epoch with up to 90\% of pruning i2h weights in RNN\_ReLU, LSTM, and GRU. 
RNN\_Tanh takes about two re-training epochs to recover after 90\% of i2h weight pruning.
Finally, as expected, this pruned model never recovers with 100\% of i2h weights pruning.

Subsequently, we prune only h2h weights of each recurrent layer in our trained base model. 
Results showed that we could safely prune about 70\% of h2h weights for RNN\_ReLU and LSTM, while 80\% of h2h weights for RNN\_Tanh and GRU, see Figure \ref{fig:h2h_prune_performance}.
Like pruning only i2h weights, models still recover after one re-training epoch with up to 90\% of pruning h2h weights.
Pruning 100\% h2h weights, RNN\_Tanh and RNN\_ReLU never recover, but GRU and LSTM still retain the original performance with just one re-training epoch.

%
%
\subsection{Random Structural Priors for Recurrent Neural Networks}
Another method than pruning to induce sparsity in a recurrent network is by applying prior structures by design.
We use random structures that are generated by converting random graphs into neural architectures, similar as in \cite{stier2019structural}.
For this, we begin with a random graph and calculate the layer indexing of each vertex.
A layer index is obtained recursively by $v \mapsto max(\{ind^{l}(s) | (s, v) \in E_{v}^{in}\} \cup \{-1\}) + 1$.
This layer indexing helps to identify the layer of a neural architecture a vertex belongs to.

\begin{figure}[tbh]
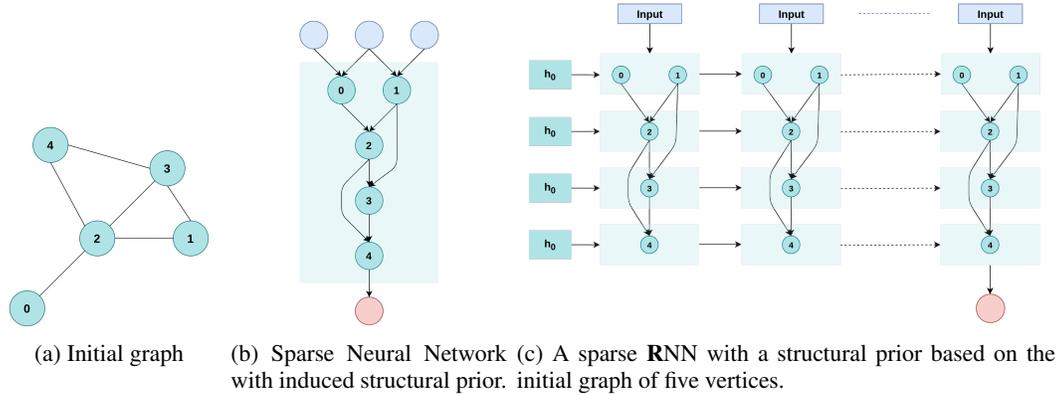

  	\vspace{\articleEnvSpaceTop}
	\centering
	\subfloat[Initial graph\label{fig:rnn_rg}]{
		\includegraphics[width=0.2\textwidth]{\visBasePath rnn_rg.png}
	}
	~
	\subfloat[Sparse Neural Network with induced structural prior.\label{fig:random_ann}]{
		\includegraphics[width=0.25\textwidth]{\visBasePath random_ann.png}
	}
	~
	\subfloat[A sparse \textbf{R}NN with a structural prior based on the initial graph of five vertices.\label{fig:recurrent_arc}]{
		\includegraphics[width=0.5\textwidth]{\visBasePath recurrent_arc.png}
	}
  	\caption{
  		We select one directed version of the graph from Figure \ref{fig:rnn_rg}, compute its topological ordering based on the described layer indexing and embed it into a neural network as a structural prior as shown in Figure \ref{fig:random_ann}.
  		This randomly structured ANN can then be converted into a randomly structured RNN by introducing recurrent connections, as in Figure \ref{fig:recurrent_arc}.
  	}
  	\label{fig:random_ann_rnn}
  	\ifthenelse{\boolean{article_dense_spacing}}{\vspace{-2em}}{\vspace{-1em}}
\end{figure}

Such a graph is used to generate randomly structured ANNs by embedding it between an input and an output layer, as in Figure \ref{fig:random_ann}.
RNNs can be understood as a sequence of neural networks, in which a network model at sequence $t$ accepts outputs from a model at sequence $t-1$.
Introducing recurrent connections as in Figure \ref{fig:recurrent_arc} provides us with Sparse RNNs with random structure.

We generate 100 connected Watts--Strogatz \cite{watts1998collective} and 100 Barabási--Albert \cite{barabasi1999emergence} graphs using the graph generators provided by \href{https://networkx.org}{NetworkX}.
The graphs are transformed into recurrent networks and trained on the \texttt{Reber} grammar dataset.
Analogue to pruning, this experiment is also conducted with RNN with Tanh nonlinearity, RNN with ReLU nonlinearity, LSTM, and GRU.
\ifthenelse{\boolean{article_extended}}{
	\autoref{fig:performance_differences} shows the performance differences between Watts-Strogatz and Barabási-Albert graphs.
	\begin{figure}[h]
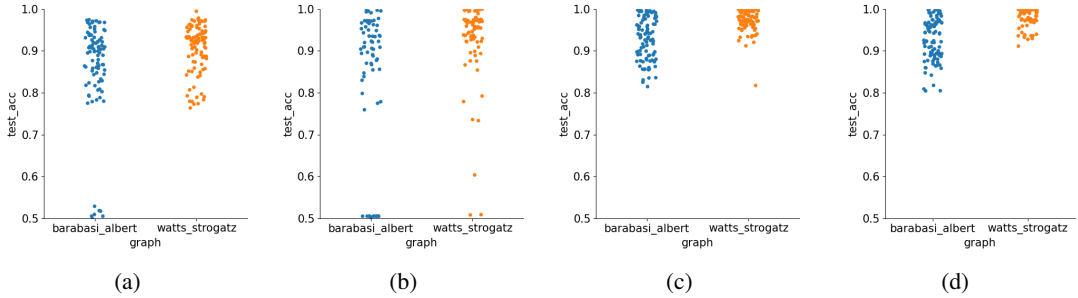

		\centering
		\subfloat[\label{fig:tanh_graph_test_acc}]{
			\includegraphics[width=0.25\textwidth]{\visBasePath tanh_graph_test_acc.png}
		}
		\subfloat[\label{fig:relu_graph_test_acc}]{
			\includegraphics[width=0.25\textwidth]{\visBasePath relu_graph_test_acc.png}
		}
		\subfloat[\label{fig:lstm_graph_test_acc}]{
			\includegraphics[width=0.25\textwidth]{\visBasePath lstm_graph_test_acc.png}
		}
		\subfloat[\label{fig:gru_graph_test_acc}]{
			\includegraphics[width=0.25\textwidth]{\visBasePath gru_graph_test_acc.png}
		}
	  
	  	\caption{
			The performance differences between Watts–Strogatz and Barabási–Albert graphs for (\ref{fig:tanh_graph_test_acc}) RNN with Tanh nonlinearity, (\ref{fig:relu_graph_test_acc}) RNN with ReLU nonlinearity, (\ref{fig:lstm_graph_test_acc}) LSTM, and (\ref{fig:gru_graph_test_acc}) GRU.
	  	}
	  	\label{fig:performance_differences}
	\end{figure}
}{}

To identify essential graph properties that correlate with the performance, we calculated the Pearson correlation of each graph property to its corresponding performance results.
\autoref{tab:pearson_corr} shows the Pearson correlation between test accuracy and different graph properties.

\begin{table}[tbh]
  	\vspace{\articleEnvSpaceTopSmall}
	\centering
	\small
	\begin{tabular}{|l|c|c|c|c|}
	    \hline
		\textbf{Property} & \multicolumn{4}{|c|}{\textbf{Correlation with test accuracy}}\\
		\hline
		 & RNN\_Tanh & RNN\_ReLU & LSTM & GRU\\
		\hline
		layers & 0.25 & 0.30 & 0.28 & 0.34\\
		nodes & \textbf{0.40} & \textbf{0.44} & \textbf{0.44} & \textbf{0.49}\\
		edges & 0.38 & \textbf{0.43} & \textbf{0.42} & \textbf{0.49}\\
		source\_nodes & 0.35 & \textbf{0.47} & \textbf{0.57} & \textbf{0.74}\\
		diameter & -0.23 & -0.27 & -0.32 & -0.20\\
		density & 0.29 & 0.15 & 0.29 & 0.34\\
		average\_shortest\_path\_length & -0.27 & -0.25 & -0.36 & -0.23\\
		eccentricity\_var & -0.22 & -0.24 & -0.30 & -0.21\\
		degree\_var & -0.28 & -0.26 & -0.39 & \textbf{-0.58}\\
		closeness\_var & \textbf{-0.46} & -0.39 & \textbf{-0.51} & \textbf{-0.67}\\
		nodes\_betweenness\_var & \textbf{-0.49} & \textbf{-0.41} & \textbf{-0.56} & \textbf{-0.52}\\
		edge\_betweenness\_var & -0.34 & -0.30 & \textbf{-0.44} & -0.26\\
		\hline
	\end{tabular}\\
	\caption{
		Pearson correlation between the test accuracy of an architecture and different graph properties.
	}
	\label{tab:pearson_corr}
	\vspace{-2em}
\end{table}

Based on this correlation, we found closeness\_var, nodes\_betweenness\_var, and the number of nodes to be essential properties for randomly structured RNN\_Tanh. 
For randomly structured RNN\_ReLU, the essential properties are the number of nodes, the number of edges, the number of source nodes, and nodes\_betweenness\_var.
In the case of randomly structured LSTM, we found six essential properties, i.e., the number of nodes, the number of edges, the number of source nodes, closeness\_var, nodes\_betweenness\_var, and edge\_betweenness\_var. 
Similarly, we found six essential properties for randomly structured GRU, namely, the number of nodes, the number of edges, the number of source nodes, degree\_var, closeness\_var, and nodes\_betweenness\_var.

By storing the graph properties and their corresponding performance during the training of randomly structured recurrent networks, we create a small dataset of 200 rows for each RNN variant. 
We then train three different regression algorithms, namely Bayesian Ridge, Random Forest, and AdaBoost, on this dataset and report an R-squared value for each.

Performance of \textbf{RNN\_TANH} was best predicted with Bayesian Ridge (BR) Regression with an $R^2$ of 0.47919, while Random Forest (RF) achieved 0.43163 and AdaBoost (AB) 0.35698.
All regressors have an $R^2$ of below 0.5, from which we conclude only a weak fit and predictability based on the used structural features.
For \textbf{RNN\_RELU}, RF was best with an $R^2$ of 0.61504, followed by AB with 0.53469 and BR with 0.36075.
Structural features on \textbf{LSTM} predicted performance with AB with 0.59514, with RF with 0.57933, and with BR with 0.37206.
We found a moderate fit for random forests, similar as in \cite{stier2019structural}.
\textbf{GRU} accuracies were predicted with AB with 0.78313 and with BR with 0.67224, and RF achieved an $R^2$ of \textbf{0.87635}.
This indicates a strong fit and a good predictability that we interpreted carefully as potentially coming from a skewed underlying distribution of the overall dataset of Sparse Neural Networks but also an indication of possible strong predictability in larger settings in which structural properties have even more impact.

%
%
\subsection{Architectural Performance Prediction in Neural Architecture Search}
We investigated a Genetic Neural Architecture Search to find correlations between structural properties of sparse priors and neural networks on a more coarse level and analysed the predictive capabilities for performances of Sparse Neural Networks when having just architectural information available.

\begin{figure}[tbh]
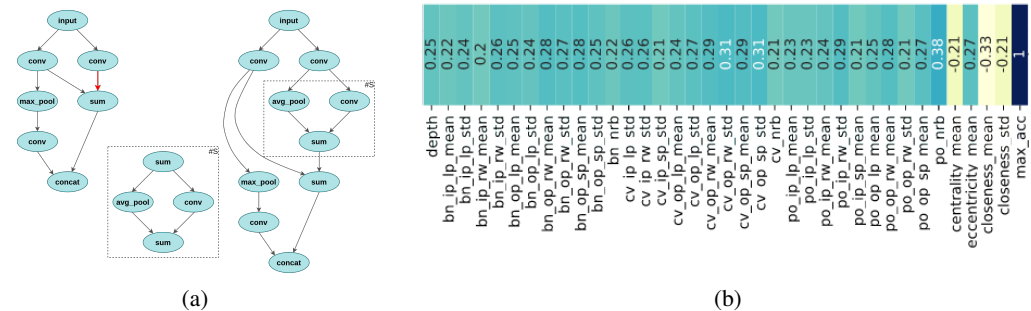

  	\vspace{\articleEnvSpaceTop}
	\centering
	\subfloat[\label{fig:genetic-nas-subgraph}]{
		\includegraphics[width=0.35\textwidth]{\visBasePath genetic-nas-subgraph-insertion.png}
	}
	~
	\subfloat[\label{fig:genetic-nas-feature_acc_corr}]{
		\includegraphics[width=0.62\textwidth]{\visBasePath genetic-nas-feature_acc_corr-reduced-horizontal.png}
	}
	\caption{Figure \ref{fig:genetic-nas-subgraph} shows an exemplary graph from our search space with a mutation through an inserted sub-graph of depth two. Figure \ref{fig:genetic-nas-feature_acc_corr} shows the correlations between structural properties and the maximum validation accuracy.}
	\vspace{-1em}
\end{figure}

\ifthenelse{\boolean{article_extended}}{
	A genetic search is a population-based search with individuals represented in a graph-based genotypical encoding such as in \autoref{fig:genetic-nas-subgraph}.
	Genotypical operations such as crossover between two genotypes distinguish the genetic from an evolutionary search.
	We employed the three operations selection, mutation, and crossover.
	
	Our analysis of the genetic search included performance prediction on purely genotypical information.
	In \cite{wendlinger2021evofficient} Wendlinger et al. investigated performance prediction for Neural Architecture Search on the search space described by Suganuma et al. in CGP-CNN \cite{suganuma2017genetic}.
	We followed a more open search space based on labelled directed acyclic graphs as in \cite{irwin2019graph} and used a genetic approach that mixes re-sampling, mutation, and cross-over instead of a purely and small $(1+\lambda)$ evolutionary approach with just point mutations.
	This has the benefit of a more diverse search space through more operations and a more loose genetic encoding scheme.
}{}

Our \textbf{search space} is based on directed acyclic graphs (DAG) and follows Irwin-Harris \cite{irwin2019graph} to represent CNN architectures as depicted in Figure \ref{fig:genetic-nas-subgraph}.
Each vertex of the DAG is labelled with an operation: a convolution, max or average pooling, a summation, or concatenation or a skip connection.
A convolution can have kernel sizes of either $3 \times 3$, $5 \times 5$ or $7 \times 7$.

In total, five \textbf{genotypical operations} were used on the search space:
two mutations and three crossover operations.
The first mutation remaps vertex operations in a genotype, e.g. it could replace \textit{max\_pool} in the left genotype of Figure \ref{fig:genetic-nas-subgraph} with an \textit{avg\_pool}.
Figure \ref{fig:genetic-nas-subgraph} depicts the result of the second mutation operation by inserting a smaller sub-graph \#S into a randomly selected edge.
The first crossover operation considers the longest simple sub-paths and swaps them between both parent genotypes.
Bridge-crossover searches for bridges in both parent genotypes and swaps the succeeding child graphs after both found bridges.
In a third crossover operation, matching layers with feature maps are searched.
The number of feature maps within all vertices of both genotypes for matching layers are averaged.

We use a minimum depth of 6 and a maximum of 12 for all DAGs.
Due to our final choice of mutation operation that increases the depth size with every generation, we set the minimum and maximum depths for the random search to 10 and 36.
The hyperparameter search uses a population size of 30, a mutation probability of 0.5, a crossover probability of 0, a probability of removing the worst candidates of 0.1, and an architectural estimation fraction of 0.5.

\paragraph{Architectural Performance Prediction}
The experiment is conducted on \texttt{cifar10} \cite{krizhevsky2009learning} and the meta dataset to investigate architectural performance prediction consists of 56 features with a total of 2,472 data points - we split it into 70\% training and 30\% testing.
The resulting meta-dataset constitutes a new supervised learning task containing graph-based features and the estimated performance of each candidate evaluated on \texttt{cifar10}.
Three categories, namely layer masses, path lengths, and remaining graph properties, make up the meta-dataset.

The layer mass is the number of channels of the current vertex times the sum of all channels of preceding vertices with an incoming connection to the current vertex.
The average and standard deviation of this layer mass are used as a graph-based feature in the meta-dataset.
For path lengths, we consider the shortest, longest and expected random walk length from a source to a target vertex.
Again, we take the average and standard deviation of these properties over all vertices in a graph and obtain 36 features over all possible vertex operations.
Further, we include the depth, edge betweenness, eccentricity, closeness, average neighborhood degree, and vertex degree as graph properties to the meta-dataset features.

Six of the ten most important features relate to standard deviations of path lengths regarding convolutional blocks or pooling layers.
When combining this result with the correlations of features and the maximum accuracy, which shows positive correlations for all these six features and the maximum accuracy, it seems that an even distribution of pooling and convolution layers benefits the performance of an architecture.
This assumption is further backed by the observation that handcrafted models like DenseNet use pooling vertices to connect architectural cells.
Mean centrality is also included in the ten most important features and shows a negative correlation to the maximum accuracy.
Centrality is an inverted measurement of the farness of one vertex to other vertices in the network.
Thus, we interpret these findings as evidence that deeper architectures perform on average better than shallower ones and that varying path lengths might support a similar effect as model ensembles.
Compare \autoref{tab:pearson_corr} and Figure \ref{fig:genetic-nas-feature_acc_corr} for correlations between graph properties and accuracy estimations across different experiments.

\section{Discussion \& Conclusion \& Future Work}\label{sec:conclusion}
We presented experimental results on different methods for optimizing structures of neural networks:
pruning, neural architecture search, and prior initialization.
These methods unite under joint questions of how structure influences the performance of neural networks given data.
Structural performance prediction is an emerging method to exploit this fact to speed up search procedures or to control bias towards desirable properties such as low memory or energy consumption or computational speed for specialized hardware.

We compared five pruning techniques for pruning feed-forward networks, namely random pruning, magnitude class blinded, magnitude class uniform, magnitude class distributed, and optimal brain damage. 
Out of these five, random pruning immediately showed an accuracy drop, while the other four performed consistently near original performance for over 60\% compression rate. 
In the end, magnitude class blinded outperformed the remaining four.

While applying pruning on recurrent networks, we found models to perform consistently better for up to 60\% pruning. 
All the pruned models regain the original performance in just one to two epochs of re-training. 
This means these recurrent networks can achieve the same results as any dense recurrent network with almost 60\% fewer weights. 
As opposed to our expectations, LSTM and GRU recovered even after 100\% pruning of hidden-to-hidden weights. 
In LSTM, this might be due to a separate cell state that acts as long-term memory.

Our experiment with random structural priors for recurrent networks aimed to find essential graph properties and use them for performance prediction. 
Similar to the results of \cite{stier2019structural}, three of the essential features are the number of edges, vertices, and source nodes. 
Although the construction and properties of Watts–Strogatz and Barabási–Albert random graphs are different, recurrent networks based on this two performed equally with RNN\_Tanh and RNN\_ReLU. 
Barabási–Albert based recurrent networks perform better than Watts–Strogatz based with LSTM and GRU.

Correlation analyses between structural properties and the performance of an untrained network reveal themselves to be difficult -- after all, the mere structure is, at first sight, and ignoring the structural dependencies on the input feature space, independent of data.
All the more promising is if such a relationship between structure of models and application domains can be found.
The idea that structures contain relevant information implies that architectural priors or search strategies over architectures can be heavily biased and influenced.
To what extend this bias takes shape is difficult to understand, and we hope to foster more research towards its impact.

\begin{ack}
Paul Häusner and Jerome Würf contributed to this work during their research at the University of Passau, and we thank them for their contributions and valuable discussions.

\end{ack}

\defbibheading{bibintoc}[\bibname]{%
  \phantomsection
  \manualmark
  \markboth{\spacedlowsmallcaps{#1}}{\spacedlowsmallcaps{#1}}%
  \addtocontents{toc}{\protect\vspace{\beforebibskip}}%
  \addcontentsline{toc}{chapter}{\tocEntry{#1}}%
  \chapter*{#1}%
}
\printbibliography 

\end{document}